\documentclass[letterpaper]{article} 
\usepackage{aaai2026}  
\usepackage{times}  
\usepackage{helvet}  
\usepackage{courier}  
\usepackage[hyphens]{url}  
\usepackage{graphicx} 
\usepackage{natbib}  
\usepackage{caption} 
\usepackage{algorithm}
\usepackage{algorithmic}
\usepackage{newunicodechar}
\newunicodechar{，}{,}
\usepackage{multirow}
\usepackage{booktabs}
\usepackage[table]{xcolor}
\usepackage{amsmath}
\usepackage{newfloat}
\usepackage{listings}
\DeclareCaptionStyle{ruled}{labelfont=normalfont,labelsep=colon,strut=off} 
\floatstyle{ruled}
\newfloat{listing}{tb}{lst}{}
\floatname{listing}{Listing}
\newcounter{stage}
\newcounter{stage2}
\title{Learner-Tailored Program Repair: A Solution Generator with \\Iterative Edit-Driven Retrieval Enhancement}
\author {
    Zhenlong Dai\textsuperscript{\rm 1,2}\thanks{Both authors contributed equally to this research.}, 
    Zhuoluo Zhao\textsuperscript{\rm 1,2}\footnotemark[1], 
    Hengning Wang\textsuperscript{\rm 1}, \\
    Xiu Tang\textsuperscript{\rm 1,2}, 
    Sai Wu\textsuperscript{\rm 1,2}, 
    Chang Yao\textsuperscript{\rm 1,2}\thanks{Co-corresponding authors.}, 
    Zhipeng Gao\textsuperscript{\rm 1}, 
    Jingyuan Chen\textsuperscript{\rm 1}\footnotemark[2]
}
\affiliations {
    \textsuperscript{\rm 1}Zhejiang University\\
    \textsuperscript{\rm 2}Hangzhou High-Tech Zone (Binjiang) Institute of Blockchain and Data Security\\
    \{zhenlongdai, zhuoluozhao, whning, tangxiu, wusai, changy, zhipeng.gao, jingyuanchen\}@zju.edu.cn
}

\usepackage{bibentry}

\newcommand{\TaskName}{LPR}
\newcommand{\DataSetName}{LPR-Bench}
\newcommand{\MethodName}{LSGen}

\begin{document}
\maketitle

\begin{abstract}
With the development of large language models (LLMs) in the field of programming, intelligent programming coaching systems have gained widespread attention.
However, most research focuses on repairing the buggy code of programming learners without providing the underlying causes of the bugs.
To address this gap, we introduce a novel task, namely \textbf{LPR} (\textbf{L}earner-Tailored \textbf{P}rogram \textbf{R}epair).
We then propose a novel and effective framework, \textbf{\textsc{\MethodName{}}} (\textbf{L}earner-Tailored \textbf{S}olution \textbf{G}enerator), to enhance program repair while offering the bug descriptions for the buggy code.
In the first stage, we utilize a repair solution retrieval framework to construct a solution retrieval database and then employ an edit-driven code retrieval approach to retrieve valuable solutions, guiding LLMs in identifying and fixing the bugs in buggy code.
In the second stage, we propose a solution-guided program repair method, which fixes the code and provides explanations under the guidance of retrieval solutions.
Moreover, we propose an Iterative Retrieval Enhancement method that utilizes evaluation results of the generated code to iteratively optimize the retrieval direction and explore more suitable repair strategies, improving performance in practical programming coaching scenarios.
The experimental results show that our approach outperforms a set of baselines by a large margin, validating the effectiveness of our framework for the newly proposed LPR task.
\end{abstract}
\begin{links}
\link{Code}{https://github.com/PandaAB/LSGen}
\end{links}

\section{Introduction}
\begin{figure}[htb!]
\centering
\includegraphics[width=\columnwidth]{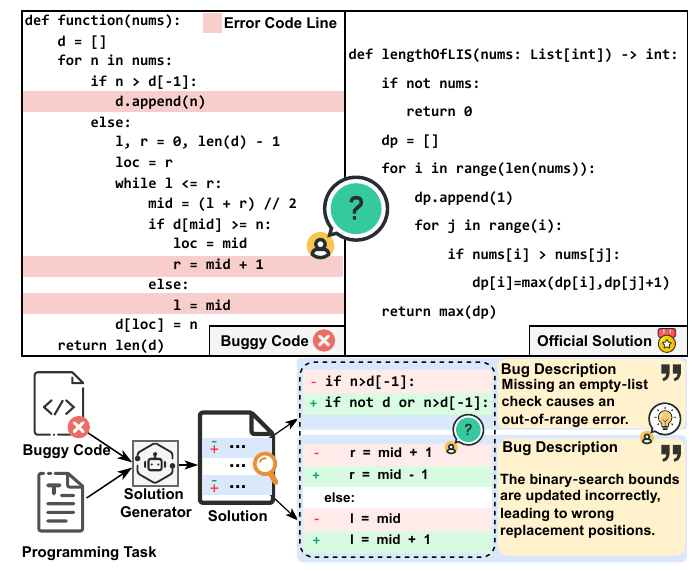}

\caption{Example of LPR. The generated solution contains the repaired code and the corresponding bug description.}
\label{fig: Introduction}

\end{figure}
As the significance of computer science and programming increases across various fields, the learning of programming has garnered widespread attention \cite{gulwani2018automated,zhang2022repairing}.
Nowadays, inspired by the promising performance of large language models (LLMs) for code~\cite{dai2024mpcoder}, researchers have applied code LLMs to intelligent tutoring for programming, utilizing them to help programming learners correct their programs.
However, most studies have focused on generating the correct patches~\cite{koutcheme2024using,koutcheme2025evaluating,phung2023generating}, and explanations for the causes of bugs are often overlooked, failing to meet the needs of actual programming learners.
Consider the following practical scenarios in Fig.~\ref{fig: Introduction}:
Alice is a programming learner who attempts to fix the bug in her code by referring to the official solution on the programming platform since she doesn't know how to resolve the issue. 
Unfortunately, she finds that the official solution is implemented differently from her code, making it difficult for her to fix her code.
The official solution adopted a dynamic programming algorithm, while she used the greedy algorithm and binary search to solve the problem of Longest Increasing Subsequence.
Therefore, Alice tried to use LLM to help her fix these bugs.
LLM provides a comprehensive and clear solution: a patch to correct the code (\textit{e.g.}, colored in green) and a bug description (\textit{e.g.}, colored in yellow) explaining why the modification is needed (\textit{e.g.}, colored in red). 
As a result, Alice can understand the changes, knowing why and how to fix her code.

There is limited research in investigating how to generate solutions that contain repair code and the corresponding bug explanation.
To address this gap, we propose a new task in this paper, namely \textbf{Learner-Tailored Program Repair}, denoted as \textbf{LPR}. This task aims to generate solutions that contain the fixed code and the corresponding bug description for the programming learner.
LPR is a non-trivial task regarding the following key is a challenging task:
(i) \textbf{Bugs in code written by programming learners are hard to identify and explain.}
Unlike the code written by professional programmers, the diverse coding styles, poor naming conventions, and chaotic implementations of programming learners make it difficult to identify and understand bugs in buggy code. 
Even after discovering the issue, understanding why the bug occurs and how it affects the program's behavior can be challenging. 
(ii) \textbf{The various and complex bugs in code written by programming learners are hard to fix.}
Due to the open-ended nature of programming problems, there are multiple possible approaches to solving the same problem. Unlike providing pre-written code, fixing bugs in different approaches demands a deeper understanding of the diverse programming patterns and problem-solving strategies employed by various users.
(iii) \textbf{Evaluating bug descriptions for buggy code is challenging}.
Unlike code correctness, which can be evaluated by executing test cases, there is no automatic evaluation metric to estimate bug descriptions. 
Evaluating the correctness of bug descriptions is
a time-consuming manual process, hindering the research about personalized programming coaching.
How to evaluate the correctness of bug descriptions quantitatively becomes another challenge for our study.

To tackle the above challenges, we propose a novel and effective framework named \textbf{\textsc{\MethodName{}}}, which is designed to generate repair code and the corresponding bug descriptions for programming users. 
First, we propose a Repair Solution Retrieval Framework to provide high-quality solution data and then leverage an edit-driven approach to obtain similar and valuable solutions for program repair.
To address the challenge of identifying and explaining bugs, 
we propose a reference-inspired solution generation approach that integrates diff analysis and textual bug descriptions from retrieval code pairs, directs LLMs to capture code modifications and their underlying causes.
To address the challenge of fixing various complex bugs in the code of programming learners, we propose an iterative retrieval enhancement method, which iteratively retrieves repair strategies that match the current generated incorrect code, thereby improving both usability and repair performance.
To address the challenge of evaluating bug descriptions, we propose an automatic evaluation metric that utilizes LLMs to achieve fine-grained logical consistency assessment.
In summary, our paper makes the following contributions:
(1) Current research mainly focuses on repairing the buggy code. To the best of our knowledge, no prior work has deeply explored how to generate the repaired code and the corresponding bug description for programming learners.
(2) We propose an automatic evaluation metric for estimating the correctness of bug descriptions.
(3) We propose a novel and effective framework, named \textsc{\MethodName{}}, that leverages the submission and evaluation system of the Programming platform to generate the repaired code and the corresponding bug description.
The experimental results show that our model outperforms a set of baselines, demonstrating its strong performance and practical usability.
We hope our study can lay the foundations for this research topic.

\section{Related Work}
Recent advancements in LLMs~\cite{dong2025knowledgepowerharnessinglarge,dong2025llm} have spurred their integration into automatic program repair in intelligent tutoring~\cite{zhang2022repairing,koutcheme2024benchmarking}.
Early work mainly focused on directly prompting LLMs to generate correct code~\cite{koutcheme2023automated,phung2024automatinghumantutorstyleprogramming}.
Cref~\cite{yang2024crefllmbasedconversationalsoftware} and TreeInstruct~\cite{kargupta2024instructassistllmbasedmultiturn} leverage multi-turn conversation to guide LLM in automatically repairing bugs.
FastFixer~\cite{liu2024fastfixerefficienteffectiveapproach} improves repair accuracy by fine-tuning.
Recent research has adopted the retrieval-augmented method to improve the accuracy of program repair.
PAR~\cite{zhao2024peer} and PyDex~\cite{zhang2024pydex} both repair code by retrieving similar correct code based on test cases.
PyFiXV~\cite{phung2023generating} performs code repair by retrieving examples based on edit distance.
Additionally, MMAPR~\cite{zhang2022repairing} and ASSIST~\cite{van2024assist} employ a hybrid of syntactic and logical repair to guide the LLM.
Existing approaches to automated program repair for learners mainly focus on generating the correct program, but neglect personalized repair needs.
Learners not only require a correct program but also seek to understand the root causes of their bugs.
Recent studies have begun to explore the capability of LLMs to generate bug descriptions for learners~\cite{koutcheme2024using}.
However, current evaluation approaches depend on manual assessment~\cite{koutcheme2025evaluating}, which is time-consuming.
To remedy this gap, we propose an automatic evaluation metric that automatically assesses the quality of generated bug descriptions.
\section{Methodology}
In this section, we introduce a generic and effective framework, \textbf{\textsc{\MethodName{}}}, aimed at enhancing program repair while providing the bug descriptions for the program learner. 
In the first stage, we utilize a repair solution retrieval framework to construct a solution retrieval database and then employ an edit-driven solution retrieval approach to retrieve valuable solutions.
In the second stage, we propose a solution-guided program repair method, 
which generates the repair solution based on diff-based program analysis and
textual bug description from candidate solutions.
Moreover, we propose an iterative repair enhancement approach that utilizes the evaluation results of the program to iteratively optimize the retrieval direction and explore more suitable repair methods for the buggy code, improving performance in practical programming learning scenarios.
\subsection{Task Definition}
Given a specific programming problem $q$, a buggy code $c$, and a collection $D$  of historical submissions from other users, the objective is to generate a solution $s$ that contains the fixed version $y$ of the buggy code $c$ along with corresponding bug descriptions $\mathcal{B}$.


\begin{figure*}[thb!]
\centering
\includegraphics[width=0.83\textwidth]{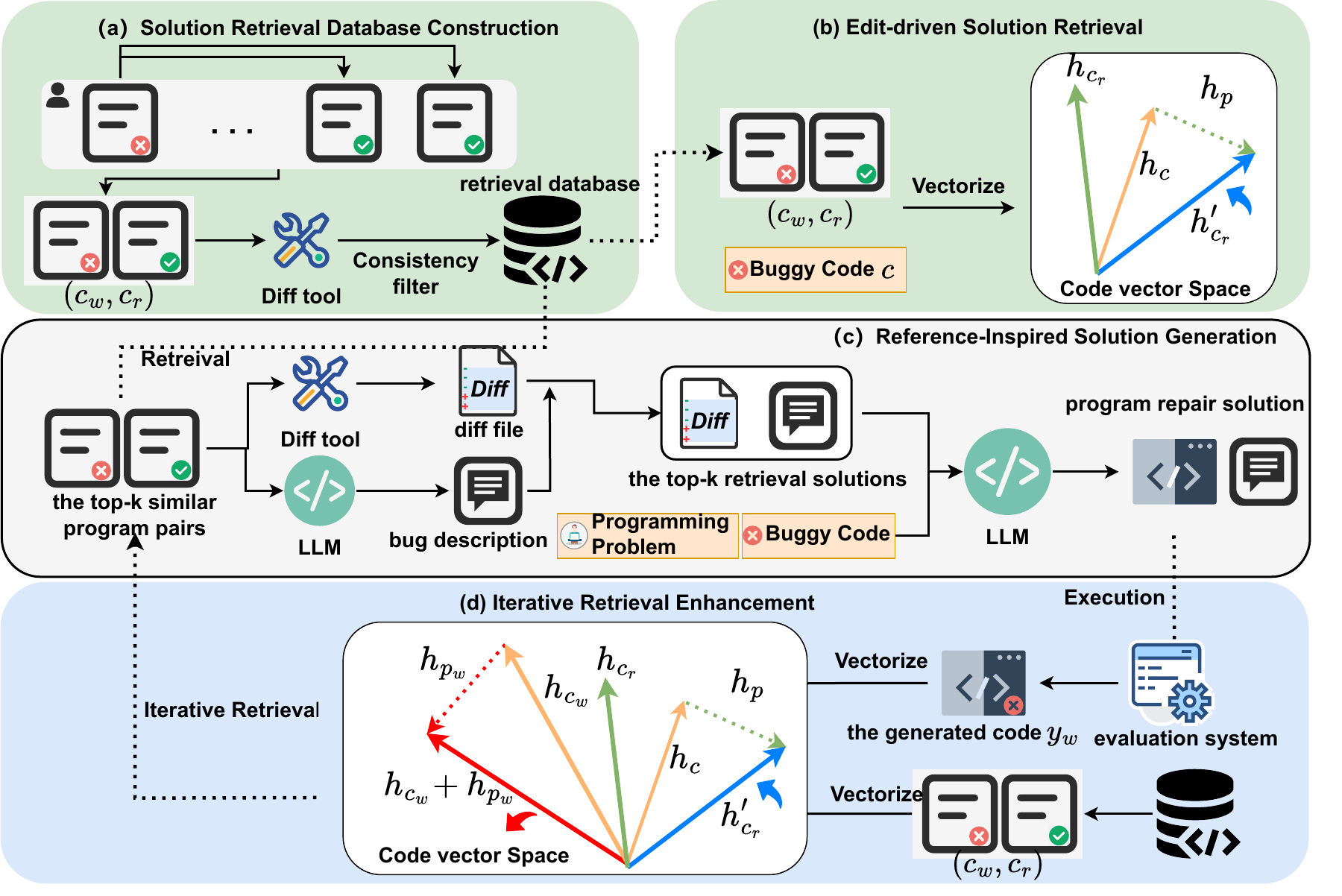}
\caption{Overview of \textsc{\MethodName{}}. (a) Illustration of the Solution Retrieval Database Construction process. (b) Illustration of Edit-driven Solution Retrieval process. (c) Illustration of the Reference-Inspired Solution Generation process.
(d) Illustration of the Iterative Retrieval Enhancement process.} 
\label{fig:model}
\end{figure*}

\subsection{Stage \Roman{stage}: Repair Solution Retrieval Framework}
Programming platforms contain a large number of submissions for the same programming problem, including incorrect submissions with bugs similar to the user's code and correct submissions that may contain potential solutions.
These submissions could provide code LLMs with valuable reference solutions to enhance the performance of program repair.
However, the open-ended nature of programming problems and differences in implementation approaches make precise and effective retrieval of solutions to buggy code challenging~\cite{wang2021novices}.
To address this challenge, we propose a repair solution retrieval framework that first constructs a solution database to provide high-quality solution data and then leverages an edit-driven approach to obtain similar and valuable solutions for program repair.

\subsubsection{Solution Retrieval Database Construction.}
To effectively retrieve valuable repair solutions for buggy code, we first construct a high-quality retrieval database from the historical submissions.
Specifically, as illustrated in Fig.~\ref{fig:model}(a), for a given programming problem $q$, we obtain a sequence of submissions for this problem from each user by sorting them based on submission time. 
For an incorrect submission, we consider all subsequent correct submissions made after the submission time as potential fixes.
Inspired by the previous study~\cite{dai2025less},  we retain only the pair with the highest consistency score that exceeds the threshold $A$ from each user's sequence of submissions.
The high-quality retrieval database is constructed as follows:
\begin{align}
D_{p} &= \{(c_w, c_r) \mid \mathop{\arg\max}_{(c_w, c_r) \in P, P \in D} F_\text{diff}(c_w, c_r) \geq A \},
\end{align}
where $D_p$ is the retrieval database, $P$ represents the code pairs of a user, each $p \in P$ consists of an incorrect and a correct code pair ($c_w$, $c_r$).
$F_\text{diff}(\cdot)$ is the function for calculating the consistency between incorrect and correct code as follows:
\begin{equation}
F_\text{diff}(c_w, c_r) = \frac{R(c_w, c_r)}{K(c_r)},
\end{equation}
where $K$ indicates the total number of code lines in the fixed code $c_r$. $R(c_w, c_r)$ indicates the number of code lines preserved in the after-modification code, which is implemented by the diff tool (\textit{i.e.}, git\footnote{\url{https://git-scm.com/}}).
\subsubsection{Edit-driven Solution Retrieval.}
The diversity of solutions to programming problems makes the implementation ideas of programs potentially different, even when the distribution of programs' test results (\textit{i.e.}, passing and failing cases) is the same.
This presents challenges in identifying similar solutions for the buggy code.
To address this challenge, we propose an edit-driven solution retrieval approach that leverages the vectorized code editing process between buggy code and its fixed version to search for repair solutions.

Specifically, as illustrated in Fig.~\ref{fig:model}(b), for a programming problem $q$, the code edit representation is achieved by vectorizing the code editing process as follows:
\begin{align}
h_{*} &= \text{CodeEncoder}(c_*), \\
h_p &= h_{c_r} - h_{c_w},
\end{align}
where $h_{c_{r}}$ and $h_{c_{w}}$ represent the vector representation of the code $c_r$ and $c_w$ through the code encoder, respectively. \( h_{p} \) is the code edit representation of a program pair $(c_r, c_w)$.
For a buggy code $c$, we first vectorize the code as $h_c$ through the code encoder.
Then we select the top-$k$ pairs with the closest vector distance between the virtual fixed version of buggy code $c$ and the correct code $c_r$ as follows:
\begin{align}
h^{\prime}_{c_r} &= h_{c} + h_{p},\\
D_k = \{(c_w, c_r) &\mid \mathop{\text{argmax}_{k}}_{(c_w, c_r) \in D_{p,q}}  F_\text{smi}(h^{\prime}_{c_r}, h_{c_{r}})\},
\end{align}
where $D_k$ denotes the top-k program repairs, $D_{p,q}$ represents the program pairs belonging to the problem $q$, $F_\text{smi}$ is the normalized cosine similarity. 
The closer the virtual fixed code $h^{\prime}_{c_r}$ is to the fixed code $c_r$, the more likely it is that codes $c$ and $c_w$ will exhibit similar bugs and repair processes. 
The edit-driven retrieval approach can identify more similar potential solutions to the buggy code, thereby offering valuable insights for program repair.

\subsection{Stage \Roman{stage2}: Solution-Guided Program Repair}
While LLMs possess strong problem-solving abilities in programming, they still face challenges in modifying others' code due to the diverse programming approaches and coding preferences of different learners \cite{Hellas_2023}.
To address this challenge, we propose a reference-inspired solution generation approach, which directs the Solution Generator to generate the repair solution based on diff-based program analysis and textual bug description.
To further improve program repair performance, we incorporate an iterative retrieval enhancement method that refines solution search and optimization using evaluation results as retrieval signals, enhancing the reliability and applicability of repairs in programming learning scenarios.
\subsubsection{Reference-Inspired Solution Generation.}
Code changes are often localized and subtle, making it difficult to track their logical evolution. 
Therefore, LLMs struggle to capture the modifications between different versions of the code, hindering the understanding of repair strategies within similar codes to fix bugs.
Additionally, the diversity of implementation methods and the distinct programming ideas of students make it very difficult to identify and understand bugs.
To address this challenge, we propose a reference-inspired solution generation approach that integrates diff analysis and textual bug descriptions.
This approach directs LLMs to capture code modifications and their underlying causes, providing fix codes along with bug explanations.
Specifically, to better understand the repair process and explain the reason for the bugs, we utilize an LLM to explain the bugs for each code pair in $D_k$, as described by the following equations:
\begin{align}
     B = \text{LLM}(q, c_w, c_r),
\end{align}
where $B$ is the textual bug description generated from the incorrect code $c_w$, providing direct insight into the causes of the bugs.
Drawing inspiration from diff tools that help developers compare and understand patterns in code modifications, as shown in Fig.~\ref{fig:model}(c), we reformat each code pair $(c_w, c_r)$ in the top-$k$ retrieval solutions into a diff file $d_{w,r}$ using a diff tool to illustrate the modification process. 
Then, we generate solutions based on the top-$k$ retrieval solutions that include the textual bug description and diff analysis as follows:
\begin{align}
     s &= \text{LLM}(q,\{z_i\}_{i=1}^{k},c),\\
     z &= (d_{w,r}, B),
\end{align}
where $s$ is the generated solution for the buggy code $c$, which contains a fixed code $y$ along with the corresponding bug descriptions $\mathcal{B}$.
$z$ is a specific solution from the retrieval dataset. 
The prompt for generation is constructed as:
\begin{figure}[!thb]
\centering
\fcolorbox{black}{gray!6}{%
\parbox{0.98\linewidth}{%
\small
\noindent$\bullet$ \textbf{Instruction}: You are a skilled programmer experienced in debugging and providing optimal code fixes. Given a programming problem and a piece of buggy code, you are required to perform the following tasks:\\
1. \textbf{Fix the Buggy Code}: fix the buggy code to meet the problem's requirements, ensuring that the changes are minimal to preserve the original structure and logic as much as possible.\\
2. \textbf{Provide Bug Descriptions}: provide clear and complete point-by-point descriptions of the bugs present in the buggy code. 
\\
\rule{\linewidth}{0.48pt} 
$\bullet$ \textbf{Programming Task}: $q$\\
$\bullet$ \textbf{The top-k program repairs for reference}: $\{d_{w_i,r_i}, B_i\}_{i=1}^{k}$\\
$\bullet$ \textbf{Buggy Code}: $c$
}%
}
\label{fig:Fault-Driven-prompt}
\end{figure}

This approach aims to effectively identify code modifications and their underlying causes in buggy code from similar solutions, ultimately enhancing automatic repair and the quality of bug descriptions.

\subsubsection{Iterative Retrieval Enhancement.}
The programming platform can evaluate the user's code through its evaluation system to determine whether the user's code is correct.
Inspired by this, we propose an iterative retrieval enhancement approach to enhance usability and repair performance in actual programming learning scenarios.
This method iteratively uses the evaluation results of the generated code as a retrieval signal and optimizes the search direction for potential solutions in the retrieval vector space.

Specifically, we define a deviation measure function $F_{w}$ to quantify the degree of deviation between the failed repair process of the generated code and the repair process of code pairs $(c_w,c_r)$ in the solution retrieval dataset $D_{p,q}$ as:
\begin{align}
    F_{w}((c,y_w),c_r) &= 1-F_{\text{sim}}(h_{c_w}+h_{p_{w}},h_{c_{r}}),\\
    h_{p_{w}} &= h_{y_w} - h_c,
\end{align}
where  $h_{p_{w}}$ is the code edit representation derived from the failed generated codes $y_w$ and the buggy code $c$. 
A larger value of $F_{w}$ indicates that the repair processes $h_{p_w}$ and $h_p$ are more different.
As shown in Fig.~\ref{fig:model}(d), we constrain the search space of the repair process through similarity and optimize the search direction based on the deviation of the generated error repair process, as described by the following equations:
\begin{equation}
\begin{split}
    F_{dis}((c,y_w), (c_w,c_r)) = \\
    F_{w}((c,y_w),c_r)&+\!F_{\text{sim}}(h_{c}\! +\! h_{p},h_{c_{r}}).
\end{split}
\end{equation}
Utilizing the evaluation results of the generated code as retrieval signals, we retrieve valuable solutions for the failed generated code \( y_w \) using the function \( F_{dis} \), and then generate solutions based on the retrieved results as follows:
\begin{align}
\hat{D}_k &= \{(c_w, c_r)\mid \mathop{\text{argmax}_{k}}_{(c_w, c_r)\in D_{p,q}}F_\text{dis}((c,y_w), (c_w,c_r))\},\\
\hat{s} &= \text{LLM}(q,\{z_i\}_{i=1}^{k},c), 
\end{align}
where \( \hat{s} \) and \( \hat{D}_k \) are the newly generated solution and retrieval dataset, and \( z_i \in \hat{D}_k \) is a solution in the re-retrieval results. 
By iterating through the above process, we continuously use the evaluation results as retrieval signals to optimize the search for solutions required for error repair, thereby improving the repair performance in real-world programming learning scenarios.

\section{Experiments}
\subsection{Experimental Setups}
\subsubsection{Benchmark.}

Existing mainstream program repair datasets for educational programming~\cite{zhao2024peer,zhang2024pydex} lack both bug descriptions of the buggy code and a database accessible for retrieval. 
To address these limitations, we introduce a benchmark, named \textbf{\DataSetName{}}, designed to evaluate the performance of Code LLMs in the \TaskName{} task. 
\DataSetName{} includes students’ buggy programs paired with their correct versions, detailed bug descriptions, and a large-scale retrieval database. 
In addition, \DataSetName{} provides an automatic evaluation framework to execute code and assess the quality of generated bug descriptions.
Our dataset is collected from the test set of ACPR~\cite{dai2025less} and further filtered by code length and the success rate of LLM-based repair to remove overly simple samples.
We construct a retrieval database containing a rich variety of user submissions, which is collected from CodeNet~\cite{puri2021codenet}.
For the data of bug descriptions, we first use GPT-4o~\cite{openai2024chatgpt4o} to generate initial bug descriptions. These generated bug descriptions are then reviewed and refined by three programmers with five years of programming experience.
The overall statistics of the dataset and the retrieval database are given in Table \ref{DatasetTable}. Further details can be found in the Appendix. 


\begin{table}[thb!]
\small
\setlength{\tabcolsep}{1mm} %
\centering
\begin{tabular}{lrrccc}
\toprule 
\multirow{2}{*}{\textbf{set}}&\multirow{2}{*}{\textbf{user}}&\multirow{2}{*}{\textbf{sample}}&\multirow{2}{*}{\textbf{bug desc}}&\multirow{2}{*}{\textbf{problem}}&\textbf{avg.problem}\\
 & & & & & \textbf{test case}\\
\midrule 
Retrieval &21,584 &274,349 & -&110 &84 \\
Test & 306 &407 & 912& 65 &89 \\
\bottomrule
\end{tabular}
\caption{Dataset Statistics.} 
\label{DatasetTable}
\end{table}

\subsubsection{Evaluation Metrics.}
To comprehensively assess the performance of a method on the \TaskName{} task, we employ the following \textbf{program evaluation metrics} to measure the quality of the generated repaired code:
(1) Code Accuracy Rate (Acc): It represents the percentage of code that successfully passes all test cases of the programming problem~\cite{muennighoff2023octopack}.
(2) Code Improvement Rate (Improve): This metric measures the average improvement rate for each piece of buggy code. It calculates the proportion of additional test cases passed after the buggy code is modified~\cite{dai2025less}.
To evaluate the quality of the generated bug descriptions, we propose an \textbf{automatic evaluation metric for bug descriptions} that determines whether the generated answers match the ground truth.
Since natural language is difficult to assess, existing studies rely on time‑consuming manual evaluation~\cite{sarsa2022automatic}.
To address this gap, we propose an automated evaluation metric that leverages LLMs to assess generated bug descriptions in a structured manner.
Formally, given a programming task $q$, a buggy code $c$, the ground truth set of the bug descriptions $\mathcal{A}=\{a_i\}_{i}^{u}$ for $c$ and the generated descriptions form the set $\mathcal{B}=\{b_i\}_{i}^{v}$. 
We ask the LLM to generate bug descriptions point by point.
The metric is defined as follows:
\begin{equation}
    M=\mathcal{K}(q, c, \mathcal{A},\mathcal{B}),
\end{equation}
where $\mathcal{K}(\cdot)$ is a function that measures the matching degree $M$ between two sets of bug descriptions. 
For a pair of bug descriptions $(a_i\in\mathcal{A}, b_j\in\mathcal{B})$, consistency is computed as follows:
\begin{equation}
    m_{i,j}=\mathcal{M}(q,c,a_i,b_j),
\end{equation}
where $\mathcal{M}(\cdot)$ is a function that uses an LLM to determine whether two bug descriptions refer to the same bug in logic, which returns 1 if they are identical and 0 otherwise.
Then we use Precision, Recall, and F1~\cite{fang2023manner} to evaluate the quality of the generated bug descriptions. 
Our evaluation metrics for bug descriptions can automatically assess generated bug descriptions at a fine-grained level, thereby reducing the comprehension burden on the LLM.
Further details can be found in the Appendix.

\begin{table*}[thb!]
\small
\setlength{\tabcolsep}{2mm}
\centering
\begin{tabular}{cllrrrrr}
\toprule
\multirow{2}{*}{\textbf{Model}}&\multirow{2}{*}{\textbf{Method}}&\multirow{2}{*}{\textbf{Retrieval Method}}&\multicolumn{2}{c}{\textbf{Program}}&\multicolumn{3}{c}{\textbf{Bug Description}}\\
&&& Acc& Improve& B-Precision& B-Recall& B-F1\\
\midrule
\multirow{7}{*}{GPT-4o} 
& NoRef &-& 18.43& 22.66& 8.90& 12.66& 9.93 \\
&AdaPatcher&-& 19.41& 23.12& 9.83& 12.35& 10.10\\
&PAR&PSM&\underline{40.54}&\underline{43.83}&20.47&\underline{29.23}&\underline{22.83}\\
&PyDex&Hamming Distance& 37.10& 41.61& \underline{21.17}& 23.33& 21.15\\
&PyFiXV&Edit Distance& 26.04& 30.38& 14.63& 18.11& 15.24\\
\cmidrule(lr){2-8}
&\textsc{\MethodName{}}\textsubscript{base}&Edit-driven Retrieval
&80.59&81.80&30.20&46.18&34.20\\
&\textsc{\MethodName{}}\textsubscript{iter3}&Iterative Retrieval Enhancement
&\textbf{91.40}&\textbf{92.11}&\textbf{33.76}&\textbf{52.38}&\textbf{38.46}\\
\midrule
\multirow{7}{*}{Claude-4} 
& NoRef &-& 18.18& 22.39& 8.68& 12.41& 9.71 \\
&AdaPatcher&-& 18.43& 22.47& 9.09& 12.55& 9.72\\
&PAR&PSM&38.08&41.67&19.96&27.55&22.04\\
&PyDex&Hamming Distance& 38.82& 42.99& 21.23& 25.02& 21.87\\
&PyFiXV&Edit Distance& \underline{42.01}& \underline{45.23}& \underline{22.78}& \underline{30.51}& \underline{24.59}\\
\cmidrule(lr){2-8}
&\textsc{\MethodName{}}\textsubscript{base}&Edit-driven Retrieval
&81.33&82.06&30.18&46.55&34.56\\
&\textsc{\MethodName{}}\textsubscript{iter3}&Iterative Retrieval Enhancement
&\textbf{91.65}&\textbf{92.02}&\textbf{33.50}&\textbf{51.55}&\textbf{38.27}\\
\midrule
\multirow{7}{*}{Qwen2.5-Coder-7B} 
& NoRef &-& 6.39& 8.61& 2.74& 3.56& 2.82 \\
&AdaPatcher&-& 8.35& 10.71& 3.81& 3.36& 3.40\\
&PAR&PSM&\underline{37.35}&\underline{38.64}&\underline{6.59}&\underline{9.88}&\underline{7.11}\\
&PyDex&Hamming Distance& 17.20& 17.88& 3.72& 4.42& 3.64\\
&PyFiXV&Edit Distance& 8.60& 10.94& 3.24& 2.93& 2.86\\
\cmidrule(lr){2-8}
&\textsc{\MethodName{}}\textsubscript{base}&Edit-driven Retrieval
&46.19&47.70&17.53&22.62&18.32\\
&\textsc{\MethodName{}}\textsubscript{iter3}&Iterative Retrieval Enhancement
&\textbf{57.49}&\textbf{58.47}&\textbf{21.62}&\textbf{27.50}&\textbf{22.48}\\
\bottomrule
\end{tabular}
\caption{
Evaluation results on the \DataSetName{}. All results in the table are reported in percentage (\(\%\)). The best method is shown in boldface, and the best among the other baselines is underlined for each metric.
}
\label{Reuslt-model}
\end{table*}
\subsubsection{Baselines.}
To evaluate the effectiveness of \textsc{\MethodName{}}, we conduct experiments on current representative models, including GPT-4o~\cite{openai2024chatgpt4o}, Claude-4\footnote{\url{https://www.anthropic.com/news/claude-4}}, CodeLlama-7B~\cite{rozière2024codellamaopenfoundation}, and Qwen2.5-Coder-7B~\cite{hui2024qwen25codertechnicalreport}.
The results of CodeLlama‑7B are provided in the Appendix due to space constraints.
We compare \textsc{\MethodName{}} to mainstream methods for program repair in programming coaching:
(1)\textbf{NoRef}: We directly prompt the LLM to generate the fixed code and corresponding bug descriptions without extra context information.
(2) \textbf{AdaPatcher}~\cite{dai2025less}
(3) \textbf{PAR}~\cite{zhao2024peer}
(4) \textbf{PyDex}~\cite{zhang2024pydex}
(5) \textbf{PyFiXV}~\cite{phung2023generating}
To evaluate the generalizability of our approach, we conduct experiments using GPT-4o on three retrieval models: Qwen3-Embedding-0.6B~\cite{zhang2025qwen3embeddingadvancingtext}, inf-retriever-v1~\cite{infly-ai_2025}, and UniXcoder~\cite{guo2022unixcoderunifiedcrossmodalpretraining}.
Further implementation details can be found in the Appendix.
\subsubsection{Implementation Details.}
For the retrieval model, we employ Qwen3-Embedding-0.6B. 
For retrieval, the number of top-$k$ solutions is set to 5. 
In our experiment, the temperature of generation is 0.2.
For the valuation of bug descriptions, we use GPT-4o-mini with the temperature set to 0.0.
Further details can be found in the Appendix.

\subsection{Experimental Results}
\subsubsection{RQ1. Effectiveness Evaluation.}
In this RQ, we aim to evaluate the effectiveness of \textsc{\MethodName{}} on the \TaskName{} task. 
Table \ref{Reuslt-model} shows the experimental results of \textsc{\MethodName{}} and the baselines on \DataSetName{}. 
We include two variants of \textsc{\MethodName{}}: \textsc{\MethodName{}}\textsubscript{base} (w/o Iterative Retrieval Enhancement) and \textsc{\MethodName{}}\textsubscript{iter3} that applies three iterations of Iterative Retrieval Enhancement. 
It is obvious that: 
(1) In terms of code accuracy, \textsc{\MethodName{}} outperforms other baselines (\textit{e.g.}, PAR, PyDex) by a large margin. 
For example, \textsc{\MethodName{}}\textsubscript{iter3} achieves 91.40\% on GPT-4o, 50.86\% higher than the second-best method, and it reaches 91.65\% on Claude-4, 49.64\% above the second-best method.
(2) In terms of bug description, \textsc{\MethodName{}} demonstrates an obvious advantage over other baselines. 
For example, \textsc{\MethodName{}}\textsubscript{iter3} achieves a B-F1 of 38.46\% on GPT-4o, 15.63\% higher than the second-best method, while it achieves 38.27\% on Claude-4, 13.68\% higher than the second-best method.
\textsc{\MethodName{}} enables the LLM to generate precise bug descriptions that reflect a deeper understanding of the underlying bugs.
Additionally, we conducted a case study to illustrate the effectiveness of \textsc{\MethodName{}}. Detailed analysis is provided in the Appendix.
\textbf{Overall, \textsc{\MethodName{}} demonstrates superior performance compared to other baselines, validating the effectiveness of our framework for the proposed \TaskName{} task.}

\begin{figure*}[thb!]
    \centering
    \includegraphics[width=1.0\textwidth]{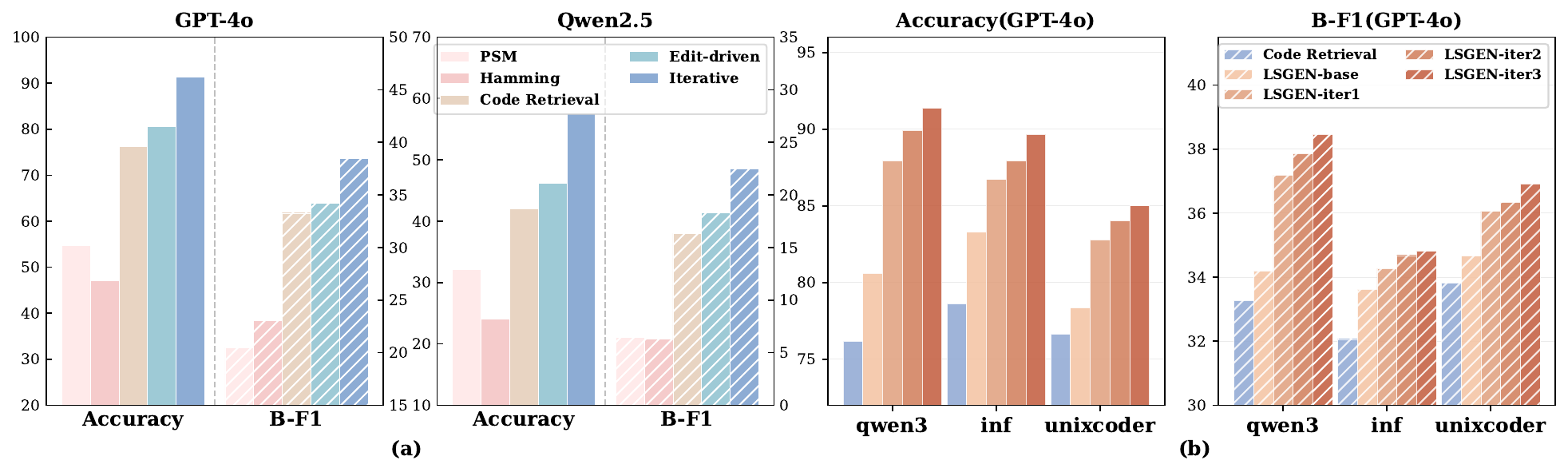}
    \caption{(a) illustrates the results of different retrieval methods on GPT‑4o and Qwen2.5‑Coder‑7B.(b) shows the effect of varying iteration counts using different retrieval models on GPT‑4o. All results in the table are reported in percentage (\%).}
    \label{fig: Retrieval Method Analysis}
\end{figure*}

\subsubsection{RQ2. Ablation Study.} 
In this RQ, we conduct an ablation study to assess the contribution of different techniques by removing each component from \textsc{\MethodName{}}. 
The specific ablation setting can be found in the Appendix.
In particular, we set the number of iterations of our iterative retrieval enhancement to 1.
The experimental results are illustrated in Table \ref{Ablation-study}. 
It is obvious that:
(1) Removing each component results in a decrease in the performance of accuracy and B-F1, which demonstrates the effectiveness of each component.
(2) Across all models, Edit-driven Solution Retrieval delivers stable improvements on both accuracy and B-F1.
(3) Removing reference-inspired solution generation results in a notable drop, particularly for smaller models (\textit{i.e.}, Qwen2.5-Coder-7B), as this component effectively identifies code modifications and their underlying causes from similar solutions, ultimately improving accuracy and B-F1.

\begin{table}[htbp]
\centering
\small
\setlength{\tabcolsep}{1.5mm}
\begin{tabular}{lccr}
\toprule
Method & Acc & Improve & B-F1 \\
\midrule
\textsc{\MethodName{}}\textsubscript{GPT-4o} & \textbf{87.96}& \textbf{88.53}&\textbf{37.19} \\
w/o Edit-driven Solution Retrieval&86.00&87.04&34.91 \\
w/o Iterative Retrieval Enhancement & 80.59&81.80&34.20\\
w/o Reference-Inspired Solution & 77.40&78.81&32.23\\
\midrule
\textsc{\MethodName{}}\textsubscript{Claude-4}  & \textbf{87.71}&\textbf{88.68}&\textbf{37.10} \\
w/o Edit-driven Solution Retrieval&86.49&87.22&34.83 \\
w/o Iterative Retrieval Enhancement &81.33&82.06&34.56 \\
w/o Reference-Inspired Solution &78.62&79.70&35.58 \\
\midrule
\textsc{\MethodName{}}\textsubscript{Qwen2.5} & \textbf{51.11}&\textbf{51.93}&\textbf{19.91} \\
w/o Edit-driven Solution Retrieval&48.65&49.44&16.96 \\
w/o Iterative Retrieval Enhancement &46.19&47.70&18.32 \\
w/o Reference-Inspired Solution &25.55&26.61&7.95 \\
\bottomrule
\end{tabular}
\caption{\label{Ablation-study}
Ablation study.
}
\end{table}

\subsubsection{RQ3. Human Study for Bug Descriptions.}
To verify the effectiveness of our metric, in this RQ, we conduct a human study to manually assess the quality of the generated bug descriptions.
We randomly select 189 samples generated by GPT-4o and pair each generated bug description with ground truth, yielding 1,390 pairs.
These pairs are provided to two experienced evaluators, who independently judge whether each generated description matches the ground truth. 
The first author then leads a discussion to resolve any disagreements.
Table \ref{Human-study} shows the results of the human study.
At the sample level, our metric is consistent with that of human evaluators in 67.72\% of cases. At the point level, this consistency is up to \textbf{93.02\%}.
Then, we calculate the Pearson correlation coefficient $r$~\cite{zhou2024divergences}, obtaining $r=0.848$, which indicates a significant linear correlation between the automatic metric and the human evaluation.
This high consistency demonstrates the effectiveness of our automatic evaluation metric.

\begin{table}[htb!]
\centering
\small
\begin{tabular}{lrrr}
\toprule
Level & Consistency & Inconsistency & Indeterminate\\
\midrule
Sample &\textbf{67.72\%} &21.16\%& 11.12\% \\
Point &\textbf{93.02\%} & 5.11\%& 1.87\%\\
\bottomrule
\end{tabular}
\caption{\label{Human-study}
Human study.
}
\end{table}
\subsubsection{RQ4. The effectiveness of Solution Retrieval.}
To validate the effectiveness of our proposed retrieval method, we conducted comparative experiments across different retrieval methods and retrieval models.
As illustrated in Fig.\ref{fig: Retrieval Method Analysis}(a), 
we compare our method with common retrieval methods used in program repair.
The iteration of our Iterative Retrieval Enhancement is set to 3.
Code Retrieval refers to retrieval based on the cosine similarity between two buggy codes.
The experimental result shows that:
\textsc{\MethodName{}} significantly outperforms other methods in both code accuracy and bug description across open-source and closed-source models. 
For example, Iterative Retrieval Enhancement is 15.23\% higher in code accuracy and 5.13\% higher in B-F1 than Code Retrieval on GPT-4o.
To evaluate the generalizability of \textsc{\MethodName{}}, we conduct experiments using three retrieval models (\textit{e.g.}, Qwen3-Embedding-0.6B, inf-retriever-v1, and UniXcoder) on both GPT‑4o and Qwen2.5‑Coder‑7B. 
As illustrated in Fig.\ref{fig: Retrieval Method Analysis}(b), we conduct experiments varying the number of iterations from 0 to 3.
Comparing \textsc{\MethodName{}}\textsubscript{iter1} with \textsc{\MethodName{}}\textsubscript{base} shows a significant improvement in the first iteration, and \textsc{\MethodName{}} continues to improve steadily as the number of iterations increases.
For example, on Qwen3-Embedding-0.6B, \textsc{\MethodName{}}\textsubscript{iter1} outperforms the \textsc{\MethodName{}}\textsubscript{base} by 7.37\% in accuracy and 2.99\% in B-F1.
From \textsc{\MethodName{}}\textsubscript{iter1} to \textsc{\MethodName{}}\textsubscript{iter3}, accuracy rises from 87.96\% to 91.40\%.
Across various retrievers, \textsc{\MethodName{}} consistently achieves notable improvements in both accuracy and B-F1, demonstrating its generalizability.
The results for Qwen2.5‑Coder‑7B are provided in the Appendix.


\section{Conclusion}
This study addresses the novel task of Learner-Tailored Program Repair, which aims to generate repaired code and corresponding bug descriptions for programming learners. 
We introduce \textsc{\MethodName{}}, an effective framework that first collects high-quality solution pairs and then uses reference-inspired solution generation to guide LLMs in capturing both code modifications and their underlying causes.
It further improves performance through iterative retrieval enhancement. 
We also propose an automatic evaluation metric that leverages LLMs to assess the quality of generated bug descriptions. 
Experiments on \DataSetName{} show that \textsc{\MethodName{}} outperforms a set of baselines by a large margin. 
We hope that our study establishes a foundation for personalized programming coaching and inspires future research on integrating repair and bug description generation for practical usability.

\section{Acknowledgements}
This work was supported by the National Natural Science Foundation of China (No.62307032, No.62507040), the “Pioneer” and “Leading Goose” R\&D Program of Zhejiang (2025C02022), the Ningbo “Yongjiang Talent Program” Youth Innovation Project (2024A-156-G) and CCF-Zhipu Large Model Innovation Fund (No.CCF-Zhipu202409).
\bibliography{aaai2026}

\section{Appendix}
\subsection{Experimental Setups}
\subsubsection{Benchmark}
We construct a benchmark for \TaskName{} (Learner-Tailored Program Repair) task, named \DataSetName{}.
\DataSetName{} includes learners’ buggy programs paired with their correct versions, detailed bug descriptions, and a large-scale retrieval database. 
Moreover, LPR-Bench provides an automatic evaluation framework to execute the generated code and assess the quality of the generated bug descriptions.
\paragraph{Dataset.}
Our dataset is collected from the test set of ACPR~\cite{dai2025less}. 
We further enhance this dataset through the following steps: 

(1) \textbf{Code Filtering}:
Firstly, we filter out samples with fewer than 10 lines of buggy code. 
Subsequently, we employ GPT-4o~\cite{openai2024chatgpt4o} to repair these codes. 
We exclude any samples whose repair success rate exceeds 1/3 over three attempts.
This filtering process ensures that the dataset contains high-quality and challenging samples.

\begin{figure}[htb!]
\centering
\includegraphics[width=\columnwidth]{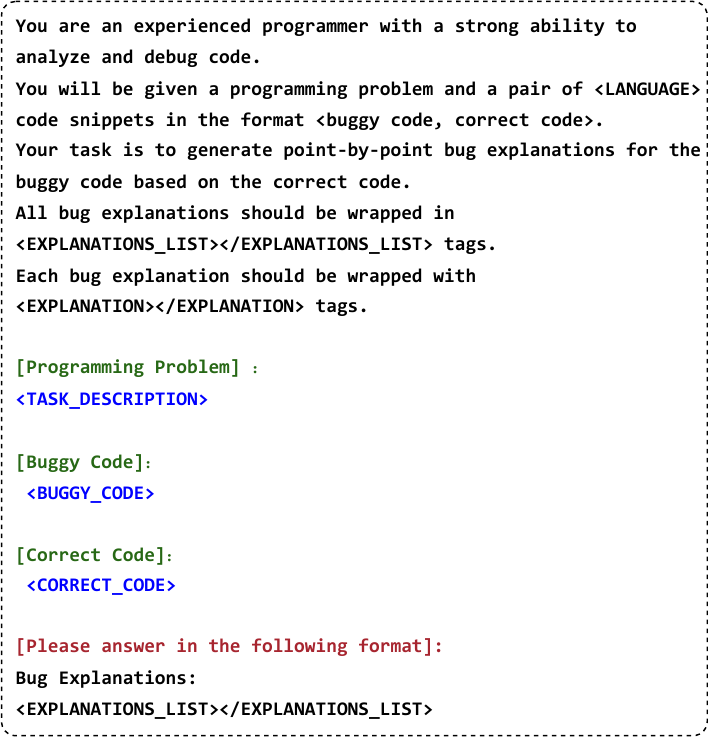}
\caption{The prompt of automated annotation.}
\label{fig: RefTextualDesc}
\end{figure}

(2) \textbf{Bug Description Annotation}:
For bug descriptions, we adopt an annotation approach that integrates automatic generation and manual refinement.
\textbf{Automatic Annotation}: As shown in Fig. \ref{fig: RefTextualDesc}, we first use GPT-4o to generate initial bug descriptions, which are written point by point. 
\textbf{Manual Annotation}: These generated bug descriptions are then reviewed and refined by three programmers with five years of programming experience. 
First, we check whether the bug descriptions generated by GPT-4o correctly identify all the bugs in the code. 
If a description is incorrect, we revise it; if any bugs are missing, we manually supplement the corresponding bug descriptions. 
Second, considering that a single bug may be resolved through multiple repair strategies, we require each annotation to focus solely on the underlying cause of the bug, without including any suggestions for how to fix it.
To ensure consistency among the three programmers, we randomly select 50 samples for manual annotation. 
Each programmer first refines the bug descriptions individually.
They subsequently compare their annotations. 
Then they agree on a unified annotation standard and apply it to the remaining samples.

As a result of these enhancement steps, we obtain 407 challenging samples. 
These samples come from 306 users across 65 programming problems, with each problem containing an average of 89 test cases.

\paragraph{Retrieval Database.}
We construct a retrieval database containing a variety of user submissions, which are collected from CodeNet~\cite{puri2021codenet}.
We filter the CodeNet dataset to include only the problem IDs listed in the ACPR and extract the associated data to construct the retrieval database.
To focus more accurately on logical changes in the code, we remove comments from the code. 
This step ensures that the retrieval process concentrates solely on the code, eliminating the influence of comments.

\paragraph{Automatic Evaluation Framework.}
To evaluate the quality of the generated solution, we employed an automated evaluation framework: 
(1) \textbf{Code Evaluation}: 
We use the tool go-judge\footnote{\url{https://github.com/criyle/go-judge}} to evaluate the generated repaired code by running it with test cases, verifying whether the produced output matches the expected results, and identifying any runtime errors.
(2) \textbf{Bug Description Evaluation}:
Since natural language is difficult to assess, existing studies rely on time-consuming manual evaluation~\cite{sarsa2022automatic}.
To address this gap, we propose an automatic evaluation metric that leverages LLMs to assess structured bug descriptions.
For more detailed information, please refer to the Evaluation Metrics section.
\subsubsection{Evaluation Metrics}
\paragraph{Program Evaluation Metrics.}
We employ the following program evaluation metrics to measure the quality of the generated repaired code: 
(1) \textbf{Code Accuracy Rate (Acc)}: It represents the percentage of code that successfully
passes all test cases of the programming problem~\cite{muennighoff2023octopack}.
The equation for calculating the code accuracy rate is as follows:
\begin{equation}
    \text{Acc}=\frac{1}{N} \sum_{i=1}^{N} \mathcal{T}(c),
\end{equation}
where $N$ is the number of samples and $c$ is the repaired code. $\mathcal{T}(\cdot)$ is a function that returns 1 if $c$ passes all test cases and 0 otherwise.
(2) \textbf{Code Improvement Rate (Improve)}: It calculates the proportion of additional test cases passed after the buggy code is modified~\cite{dai2025less}. 
The calculation equation for the code improvement rate of the $i$-th fixed code follows:
\begin{equation}
I_i = \frac{\chi (\mathcal{A})\times n}{m},     
\end{equation}
where $\chi(\cdot)$ is an indicator function that returns 1 if the condition inside the parentheses is true, and 0 otherwise. 
$\mathcal{A}$ is true if all previously passing test cases still pass after the code modification, and false otherwise.
$n$ denotes the number of cases that additional pass after repair, and $m$ represents the number of test cases that failed previously. 
The value of the $i$-th fixed code is $I_i$ if the code passes all test cases passed by the buggy code, and 0 otherwise.

\paragraph{Automatic Evaluation Metric For Bug Descriptions.}
To evaluate the quality of generated bug descriptions, we propose an automatic evaluation metric.
This metric quantifies the degree of match between the generated descriptions and the ground truth.
Formally, given a programming task $q$, a buggy code $c$, the ground truth set of the bug descriptions $\mathcal{A}=\{a_i\}_{i}^{u}$ for $c$ and the generated descriptions form the set $\mathcal{B}=\{b_i\}_{i}^{v}$. 
We ask the LLM to generate bug descriptions point by point.
The metric is defined as follows:
\begin{equation}
    M=\mathcal{K}(q, c, \mathcal{A},\mathcal{B}),
\end{equation}
where $\mathcal{K}(\cdot)$ is a function that measures the matching degree $M$ between two sets of bug descriptions. 
For a pair of bug descriptions $(a_i\in\mathcal{A}, b_j\in\mathcal{B})$, consistency is computed as follows:
\begin{equation}
    m_{i,j}=\mathcal{M}(q,c,a_i,b_j),
\end{equation}
where $\mathcal{M}(\cdot)$ is a function that uses an LLM to determine whether two bug descriptions refer to the same bug in logic, which returns 1 if they are identical and 0 otherwise.
Then we use Precision, Recall, and F1~\cite{fang2023manner} to evaluate the quality of the generated bug descriptions. 
These metrics are defined as follows:
\begin{align}
     \text{B-Precision} &= \frac{1}{N} \sum_{i=1}^{N} \frac{\text{TP}_i}{\text{TP}_i + \text{FP}_i},\\
     \text{B-Recall} &= \frac{1}{N} \sum_{i=1}^{N} \frac{\text{TP}_i}{\text{TP}_i + \text{FN}_i},\\
     \text{B-F1} &= \frac{1}{N} \sum_{i=1}^{N} \frac{2 \times \text{B-Precision} \times \text{B-Recall}}{\text{B-Precision} + \text{B-Recall}},
\end{align}
where $N$ is the number of samples; $\text{TP}_i$ denotes the number of bug descriptions correctly identified in the $i$‑th sample; $\text{FP}_i$ denotes the number of generated descriptions in the $i$‑th sample that were falsely identified as bugs; and $\text{FN}_i$ denotes the number of bug descriptions in the $i$‑th sample that the LLM failed to identify.
These metrics are defined as follows:

In real-world learning scenarios, learners require a reliable solution to correct their mistakes. 
Bug descriptions derived from repaired code that fails to pass all test cases are generally unreliable, as they may not accurately reflect the issues in the code. 
Therefore, only the bug descriptions corresponding to the fix code that passes all test points will be counted.

\subsubsection{Baselines}
\begin{figure}[htb!]
\centering
\includegraphics[width=\columnwidth]{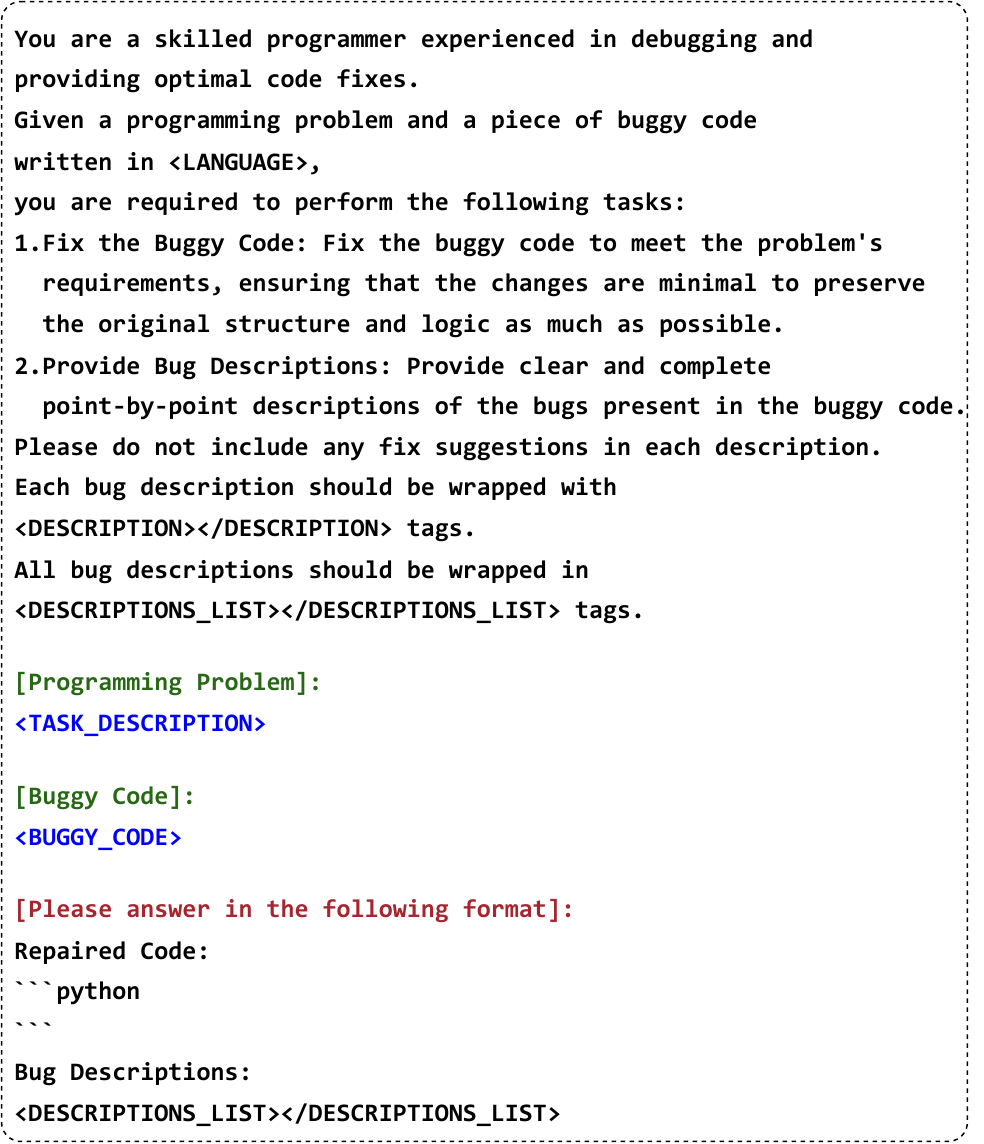}
\caption{The prompt of NoRef.}
\label{fig: NoRefPrompt}
\end{figure}

\begin{figure}[htb!]
\centering
\includegraphics[width=\columnwidth]{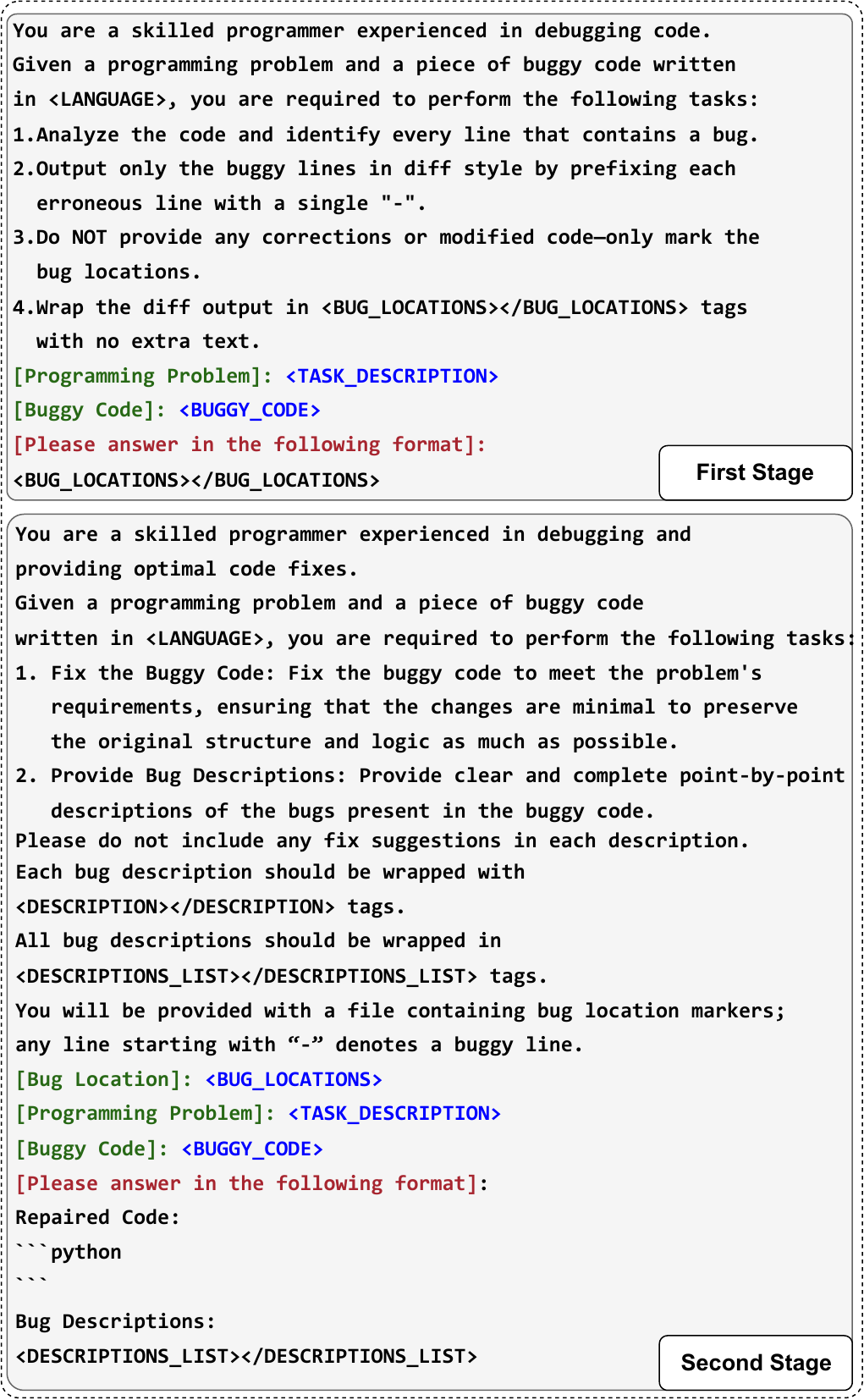}
\caption{The prompt of AdaPatcher.}
\label{fig: AdaPatcherPrompt}
\end{figure}

\begin{figure}[htb!]
\centering
\includegraphics[width=\columnwidth]{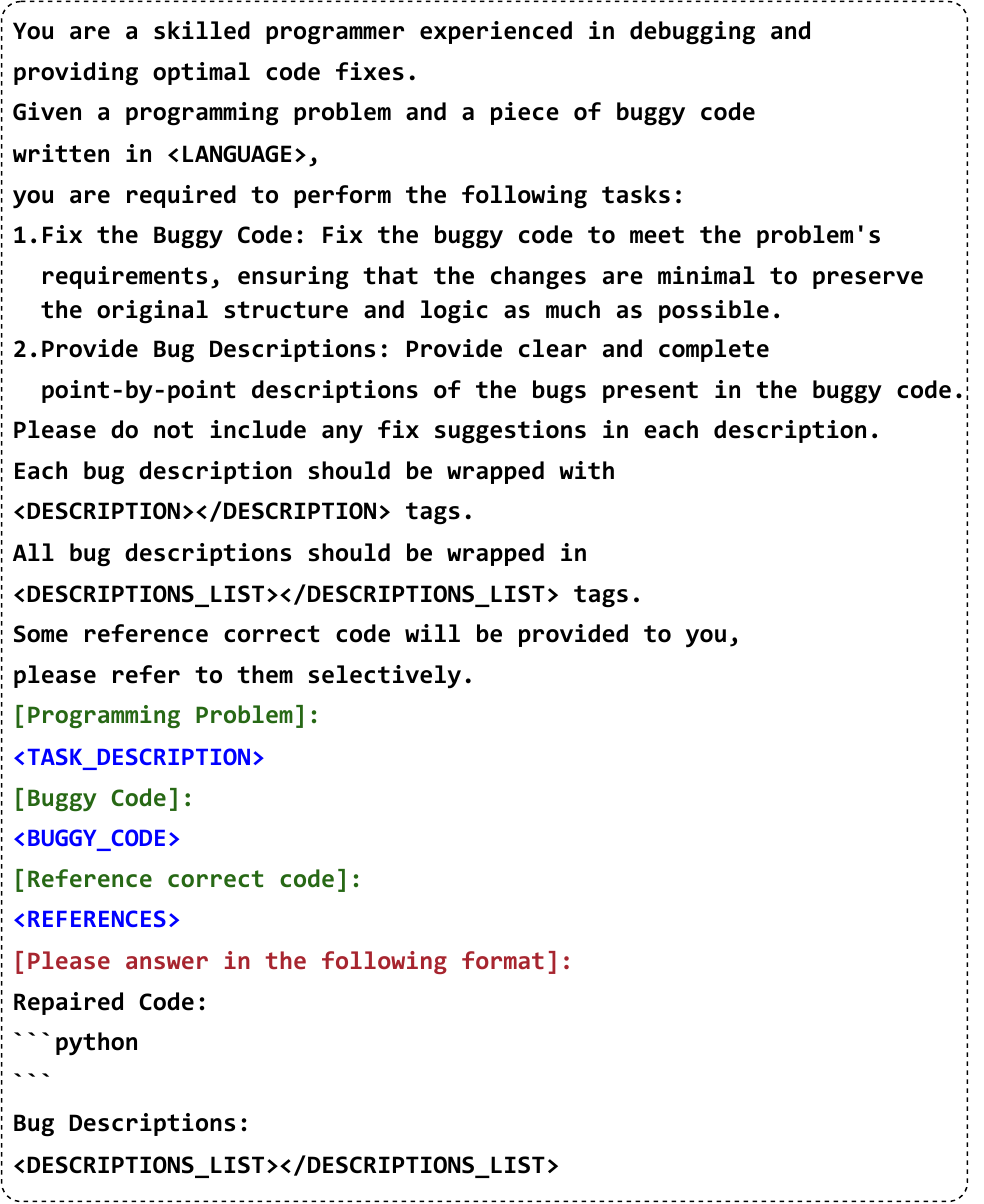}
\caption{The prompt of PAR.}
\label{fig: PARPrompt}
\end{figure}

\begin{figure}[htb!]
\centering
\includegraphics[width=\columnwidth]{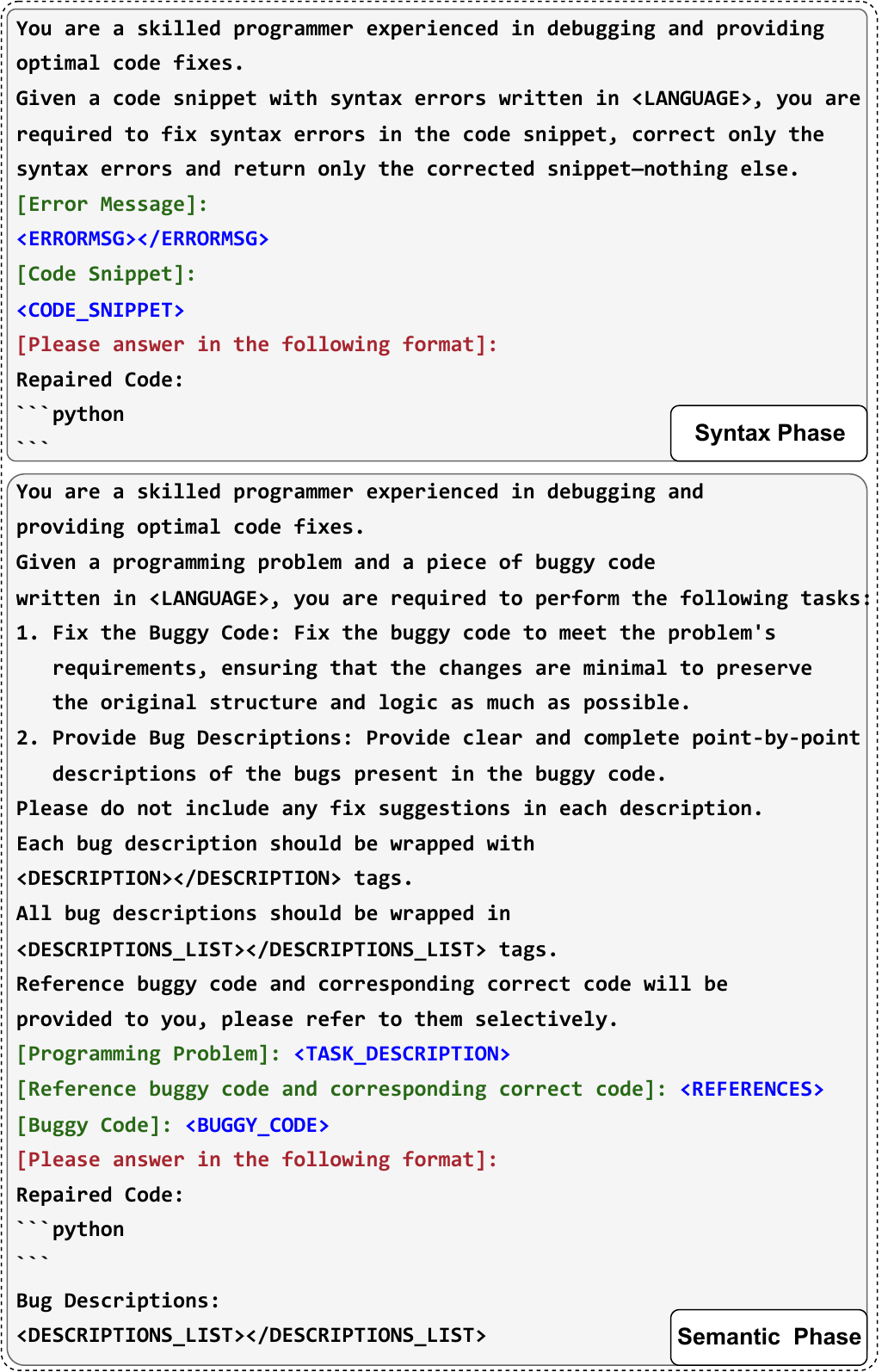}
\caption{The prompt of PyDex.}
\label{fig: PyDexPrompt}
\end{figure}

\begin{figure}[htb!]
\centering
\includegraphics[width=\columnwidth]{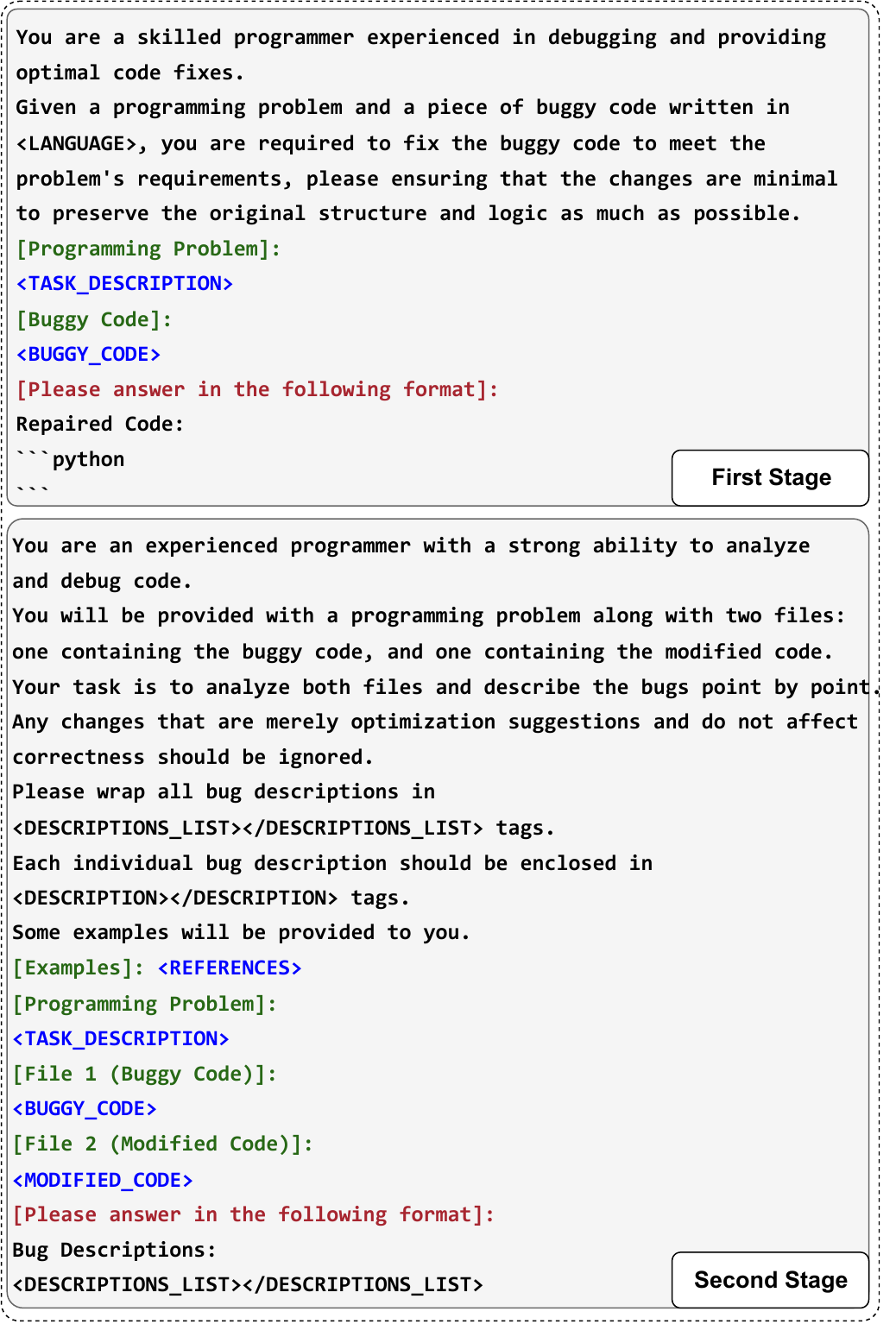}
\caption{The prompt of PyFiXV.}
\label{fig: PyFiXVPrompt}
\end{figure}

In this section, we describe the implementation details of each baseline and design specific prompts for the LRP task: 
(1) \textbf{NoRef}: As shown in Fig. \ref{fig: NoRefPrompt}, we directly prompt the LLM to generate the fixed code and corresponding bug descriptions without extra context information. In the prompt template, the \textless LANGUAGE\textgreater~ is filled with the target programming language, \textless TASK\_DESCRIPTION\textgreater~ with the problem description, and \textless BUGGY\_CODE\textgreater~ with the buggy code.
(2) \textbf{AdaPatcher}~\cite{dai2025less}: 
As shown in Fig. \ref{fig: AdaPatcherPrompt}, in the first stage, we prompt the LLM to perform bug localization on the buggy code. Then, in the second stage, we prompt the LLM to generate the repaired code along with its corresponding bug descriptions based on the localization results.
(3) \textbf{PAR}~\cite{zhao2024peer}: 
PAR repairs code by retrieving similar examples based on Peer Solution Match (PSM), which contains the measurement of multiple dimensions (\textit{e.g.}, test cases, data flow, abstract syntax tree, and BM25 score).
We assign equal weights to each of the four dimensions of PSM.
Then, as shown in Fig. \ref{fig: PARPrompt}, we provide the top-5 retrieved passing code to the LLM to generate repaired code and corresponding bug descriptions.
(4) \textbf{PyDex}~\cite{zhang2024pydex}:
PyDex repairs code in two stages, syntactic and semantic. 
As shown in Fig. \ref{fig: PyDexPrompt}: 
In the syntactic stage, Program Chunker~\cite{zhang2024pydex} extracts the code snippet containing syntax errors along with the compiler error messages, and supplies them to the LLM for correction. 
In the semantic stage, Hamming distance~\cite{singh2022improving} is used to compare the test pass and fail patterns of the buggy code with those of buggy code in the retrieval database. 
The top-5 solutions are selected, and the corresponding buggy–passing code pairs are filled into the \textless REFERENCES\textgreater ~ of the prompt template for semantic repair.
\begin{figure}[tbh!]
\centering
\includegraphics[width=0.95\columnwidth]{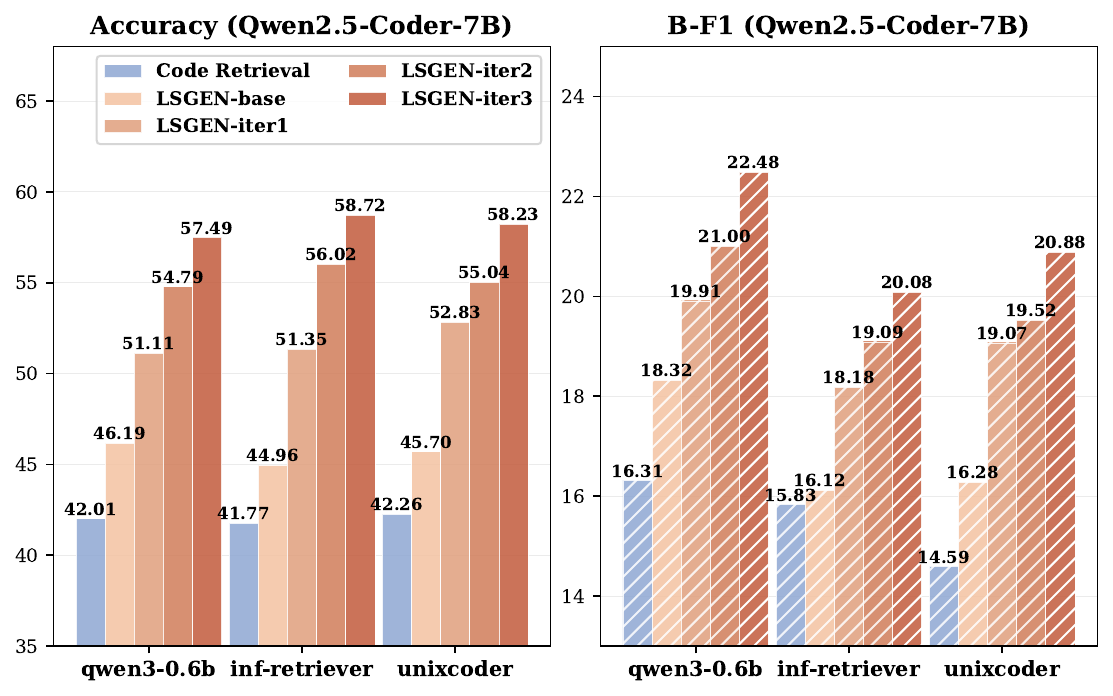}
\caption{The effect of iteration counts using different retrieval models on Qwen2.5-Coder-7B. All results in the table are reported in percentage (\%).}
\label{fig: Qwen2.5}
\end{figure}
(5) \textbf{PyFiXV}~\cite{phung2023generating}:
PyFiXV proceeds in two phases. 
As shown in Fig.\ref{fig: PyFiXVPrompt}
In the first phase, it prompts LLM to generate repaired code for each buggy code and computes the diff against the original code. 
In the second phase, it calculates the Levenshtein~\cite{rocamora2025certifiedrobustnessboundedlevenshtein} distance between this diff and the diffs of buggy–passed code pairs in the retrieval database, selects the five closest matches as reference solutions, and uses them to prompt the LLM to produce corresponding bug explanations.

\subsubsection{Implementation Details}
We employ Qwen3-Embedding-0.6B~\cite{zhang2025qwen3embeddingadvancingtext} to encode code in our retrieval pipeline, retrieving the top-5 candidate solutions. 
We generate repaired code and corresponding bug descriptions with a temperature of 0.2, and we evaluate bug descriptions using GPT-4o-mini while setting its temperature at 0.0. 
Inference for all open-source models is performed on two NVIDIA A800 GPUs(80 GB).
\paragraph{Solution Retrieval Database Construction.}
For each incorrect submission, we retain only the subsequent correct submission with the highest consistency score above the threshold $A$. 
The threshold $A$ is set to 0.65 in this paper.
\paragraph{Reference-Inspired Solution Generation.}
\begin{figure}[htb!]
\centering
\includegraphics[width=0.95\columnwidth]{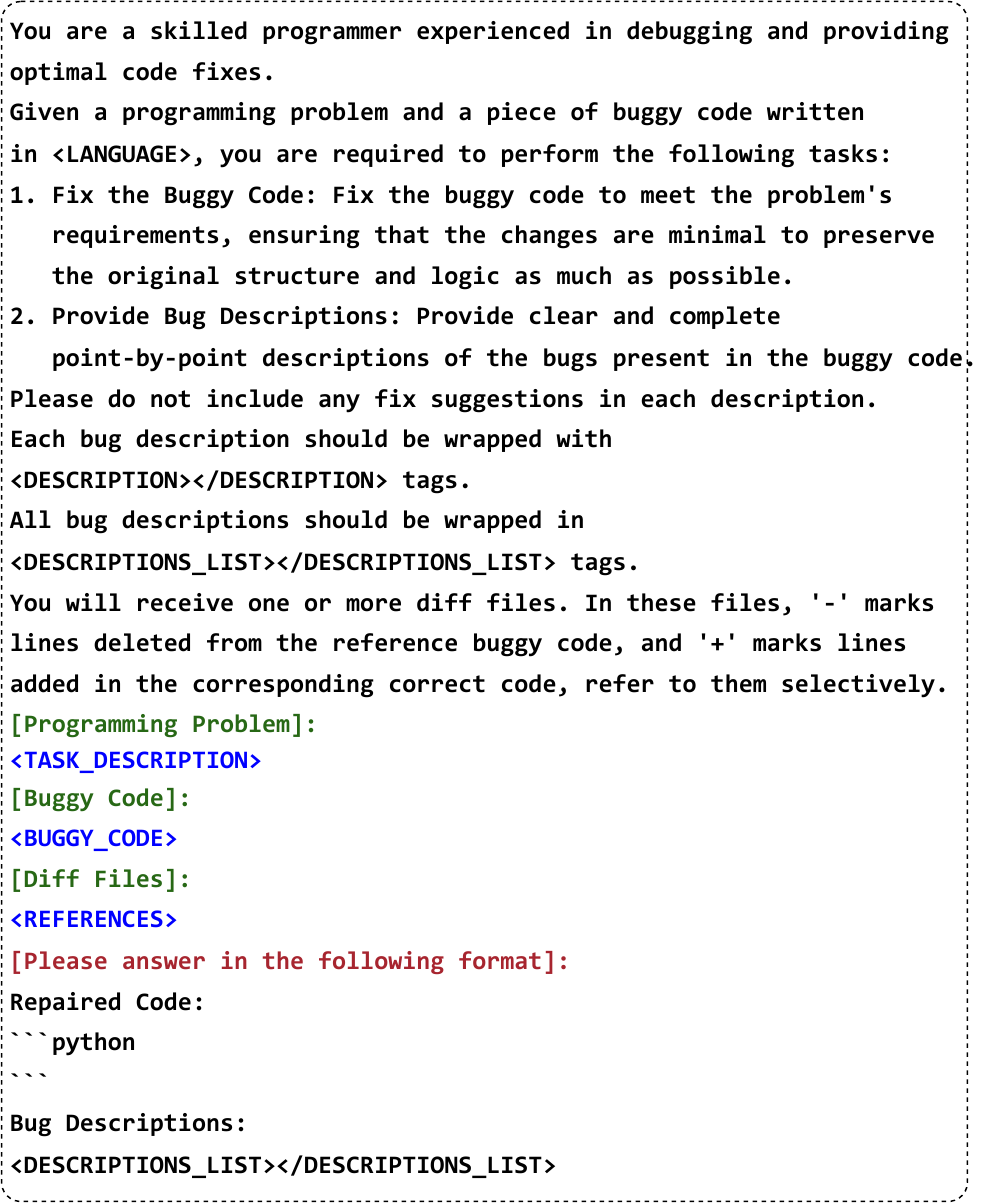}
\caption{The prompt of \textsc{\MethodName{}}.}
\label{fig: LSGenPrompt}
\end{figure}

We retrieve the top-$k$ solutions and process them in two steps. 
First, we use the prompt shown in Fig. \ref{fig: RefTextualDesc} to generate a textual bug description for each solution. 
Second, we use the diff\footnote{\url{https://git-scm.com/}} tool to compare the retrieved buggy code with its corresponding passing code, producing a diff analysis.
As shown in Fig. \ref{fig: LSGenPrompt}, we then combine the textual bug description as comments alongside the diff output and feed this combined context to the LLM, which produces both the repaired code and its corresponding bug descriptions.

\begin{table*}[thb!]
\small
\setlength{\tabcolsep}{2mm}
\centering
\begin{tabular}{cllrrrrr}
\toprule
\multirow{2}{*}{\textbf{Model}}&\multirow{2}{*}{\textbf{Method}}&\multirow{2}{*}{\textbf{Retrieval Method}}&\multicolumn{2}{c}{\textbf{Program}}&\multicolumn{3}{c}{\textbf{Bug Description}}\\
&&& Acc& Improve& B-Precision& B-Recall& B-F1\\
\midrule
\multirow{7}{*}{CodeLlama-7B} 
& NoRef &-& 4.18& 4.78& \underline{0.43}& 0.86& \underline{0.52} \\
&AdaPatcher&-&2.70& 3.33& 0.38& 0.98& 0.46\\
&PAR&PSM&22.60&23.32&0.19&2.17&0.32\\
&PyDex&Hamming Distance& \underline{28.75}& \underline{29.77}& 0.39& \underline{2.69}& \underline{0.52}\\
&PyFiXV&Edit Distance& 4.42& 5.92& 0.13& 0.49& 0.19\\
\cmidrule(lr){2-8}
&\textsc{\MethodName{}}\textsubscript{base}&Edit-driven Retrieval
&55.28&56.24&6.72&19.97&8.58\\
&\textsc{\MethodName{}}\textsubscript{iter3}&Iterative Retrieval Enhancement
&\textbf{64.86}&\textbf{65.72}&\textbf{7.95}&\textbf{23.60}&\textbf{10.15}\\
\bottomrule
\end{tabular}

\caption{
Evaluation results of the CodeLlama-7B on the \DataSetName{}. All results in the table are reported in percentage (\(\%\)). The best method is shown in boldface, and the best among the other baselines is underlined for each metric.
}
\label{Reuslt-CodeLlama}
\end{table*}
\subsection{Experimental Results}
\subsubsection{RQ1. Effectiveness Evaluation.}
\paragraph{Results of CodeLlama-7B.}
Table \ref{Reuslt-CodeLlama} presents our results on CodeLlama-7B and shows that \textsc{\MethodName{}} significantly outperforms all baselines. For example, \textsc{\MethodName{}}\textsubscript{iter3} achieves 64.86\% accuracy, which is 36.11\% higher than the second best baseline at 28.75\% (\textit{i.e.}, PyDex), and achieves B-F1 of 10.15\%, 9.63\% higher than the second best baseline.

\paragraph{Case Analysis.}
To demonstrate the effectiveness of the \textsc{\MethodName{}} framework, we conduct a case study to illustrate the \MethodName{}'s problem-solving process and how it works to fix bugs. 

\begin{figure*}[htbp] 
    \centering
    \includegraphics[width=1.0\textwidth]{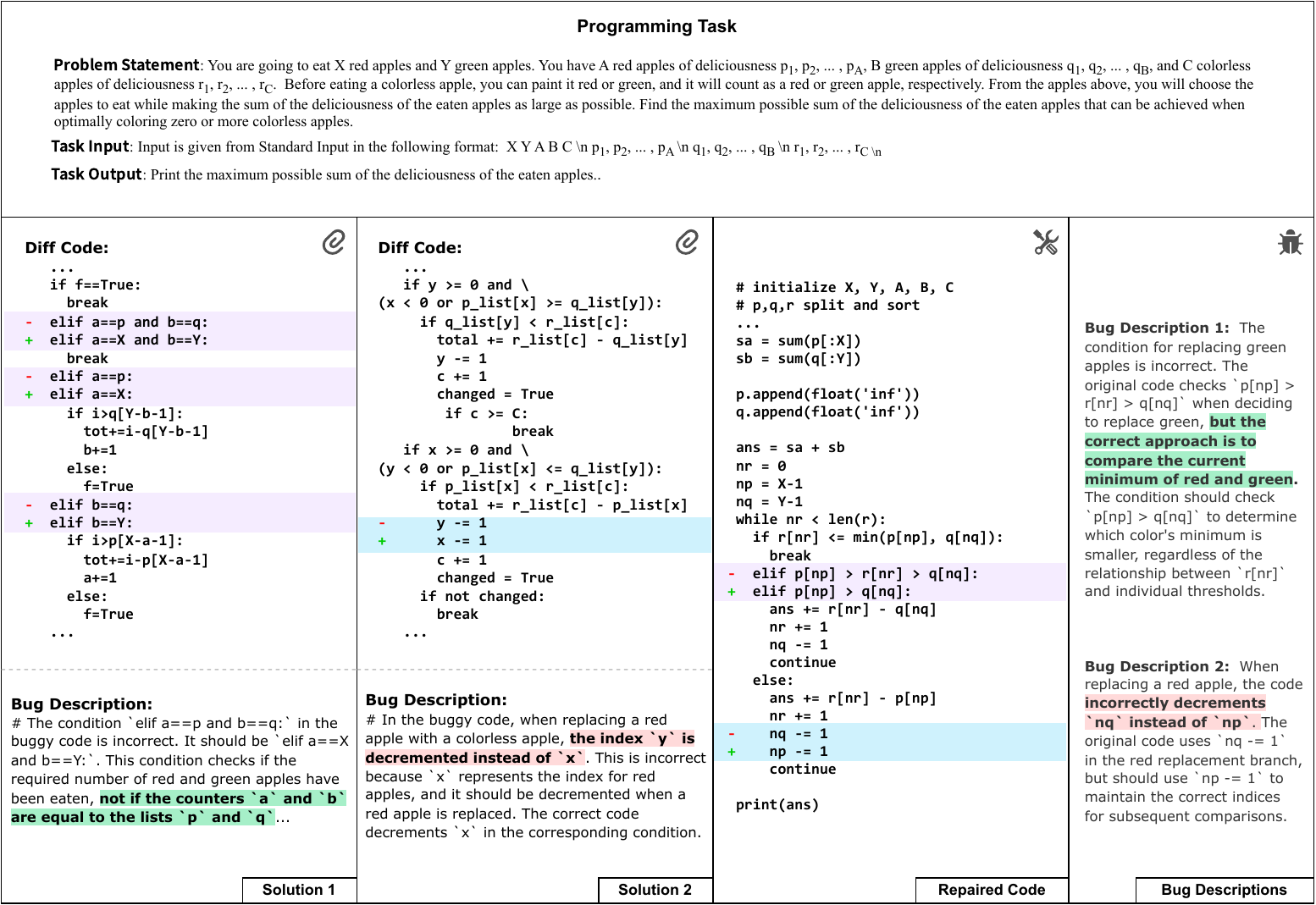}
    \caption{\textsc{\MethodName{}} correctly fixed the code by referring to the retrieval results}
    \label{fig:case_study}
\end{figure*}

Our \textsc{\MethodName{}} framework enables LLMs to obtain useful information from the solutions and generate accurate solutions.
For example, as shown in Fig.~\ref{fig:case_study}, in the buggy code, the conditional statement ``elif p[np]$>$r[nr]$>$q[nq]:'' (\textit{i.e.}, the purple block in column 3) is intended to select the minimum deliciousness apple, but it compares the wrong variables.
Notably, ``elif a==p and b==q'' in solution 1 (\textit{i.e.}, the purple block in column 1) also compares incorrect variables, leading to the wrong answer. 
By providing similar solutions (\textit{i.e.}, the purple block in column 1), \textsc{\MethodName{}} can understand the causes of bugs from bug descriptions and fix the code based on the corresponding repair process.
Likewise, solution 2 and buggy code also have a similar bug (\textit{i.e.}, the blue blocks), which results in an incorrect counter calculation.
The LLM obtains the corresponding repair process from solution 2 and performs the repair.

Ultimately, through this problem-solving process, the \textsc{\MethodName{}} framework successfully repairs the code and provides the corresponding bug descriptions, demonstrating its effectiveness.


\subsubsection{RQ2. Ablation Study Settings.}
To mitigate the impact of multiple iterations on the performance of other components, we limit the iteration of Iterative Retrieval Enhancement to 1.
We keep all other components unchanged and compute cosine similarity between the buggy code and each passed code in the retrieval database, retrieving the top-$k$ solutions. 
We refer to this variant as w/o Edit-Driven Solution Retrieval.
In the w/o Reference-Inspired Solution variant, we replace the textual bug descriptions and diff-based analysis with the retrieved solutions’ buggy and passing code.

\subsubsection{RQ4. The effectiveness of Solution Retrieval.}
To validate the effectiveness of our proposed retrieval method, we conducted comparative experiments across retrieval models on Qwen2.5-Coder-7B~\cite{hui2024qwen25codertechnicalreport}.
As shown in Fig. \ref{fig: Qwen2.5}, we conduct experiments on three retrieval models(\textit{e.g.}, Qwen3-Embedding-0.6B, inf-retriever-v1,  UniXcoder).
The number of iterations is varied from 0 to 3 in our iterative retrieval enhancement.
Comparing \textsc{\MethodName{}}\textsubscript{iter1} with \textsc{\MethodName{}}\textsubscript{base} shows a significant improvement in the first iteration, and \textsc{\MethodName{}} continues to improve steadily as the number of iterations increases.
For example, on UniXcoder, \textsc{\MethodName{}}\textsubscript{iter1} outperforms the \textsc{\MethodName{}}\textsubscript{base} by 7.13\% in accuracy and 2.79\% in B-F1.
From \textsc{\MethodName{}}\textsubscript{iter1} to \textsc{\MethodName{}}\textsubscript{iter3}, accuracy rises from 52.83\% to 58.23\%.
Across different retrievers, \textsc{\MethodName{}}~consistently achieves notable improvements in both accuracy and B-F1, demonstrating its generalizability.
\end{document}